# Real-Time Traffic Signal Control for Modern Roundabouts by Using Particle Swarm Optimization-Based Fuzzy Controller

Yue-Jiao Gong and Jun Zhang
Department of Computer Science, Sun Yat-sen University,
Guangzhou 510006, P.R. China
gongyuejiao@gmail.com

*Abstract*—Due to that the existing traffic facilities can hardly be extended, developing traffic signal control methods is the most important way to improve the traffic efficiency of modern roundabouts. This paper proposes a novel traffic signal controller with two fuzzy layers for signalizing the roundabout. The outer layer of the controller computes urgency degrees of all the phase subsets and then activates the most urgent subset. This mechanism helps to instantly respond to the current traffic condition of the roundabout so as to improve real-timeness. The inner layer of the controller computes extension time of the current phase. If the extension value is larger than a threshold value, the current phase is maintained; otherwise the next phase in the running phase subset (selected by the outer layer) is activated. The inner layer adopts well-designed phase sequences, which helps to smooth the traffic flows and to avoid traffic jam. In general, the proposed traffic signal controller is capable of improving real-timeness as well as reducing traffic congestion. Moreover, an offline particle swarm optimization (PSO) algorithm is developed to optimize the membership functions adopted in the proposed controller. By using optimal membership functions, the performance of the controller can be further improved. Simulation results demonstrate that the proposed controller outperforms previous traffic signal controllers in terms of improving the traffic efficiency of modern roundabouts.

*Index Terms*—Fuzzy logic, membership function, particle swarm optimization (PSO), roundabout, traffic control.

## I. Introduction

WITH the rapid development of automobile industry and the increase in urban population, traffic congestion is becoming a critical problem in large urban cities all around the world. As existing traffic facilities can hardly be extended due to the cost and environmental issues, traffic engineers are paying more attention on how to better use the available facilities so as to provide better service and prevent congestion. Developing traffic signal control methods is one of the most important ways to reduce congestion, to minimize vehicle delays, and to improve safety. For years, traffic signal control has been a crucial issue in intelligent transportation systems (ITS) and attracted a lot of attention [1]-[4].

Previous works done on the problem of traffic signal control are composed of fixed-time control methods and real-time control methods. Fixed-time control methods employ preset phase sequences and timings to control the traffic signals. The most famous fixed-time control method is Webster's equation for the optimal cycle length and phase timing [5]. Later, some other fixed-time signal controllers have been proposed based on Webster's equation [6]-[9]. Recently, genetic algorithms (GAs) have been widely used for fixed-time traffic signal control [4][10][11]. These stochastic optimization methods required more computational costs, but could provide better performance than the previous methods. However, the fixed-time control methods make an assumption that the traffic condition is steady. The assumption may not be valid considering that in practical the traffic flow rate changes frequently depending on the time of a day and the season of a year.

With the spread of inexpensive sensors and communication devices, real-time traffic data can be easily accessed. These years, researchers have made great efforts to develop real-time traffic signal controllers [12]-[21]. The vehicle actuated method [12] was the first real-time traffic controller. It decided whether to extend the current phase according to whether there were vehicles detecting in real time. The method was simple, and it was effective when the traffic condition was not heavy. For further study, real-time traffic control based on fuzzy logic has become a very hot research field [13]-[19], because it has been known that fuzzy logic is well suited to control complex systems with uncertainties and human perception. Basically, according to the control mechanisms, there were two types of fuzzy traffic signal controllers in use. The first type output an extension time of the current green phase [13]-[15]. If the extension time was bigger than a predefined threshold value, the current green phase would last another period of time. Otherwise, it would turn to the next phase in the phase circle. We term this fuzzy control method as FUZZY-TURN. The other type output the urgency degrees of all the phases at set intervals [16]-[18]. The phase would jump to the phase which was in the most urgent by the end of each interval. This fuzzy control method can be termed FUZZY-JUMP. Moreover, in order to improve the performance of the fuzzy systems, some researchers proposed to use genetic algorithms to optimize the membership functions used in the fuzzy systems [13][18][19]. Besides, Srinivasan et al. [3], Choy et al. [20], and Balaji and Srinivasan [21] proposed multi-agent systems which concerned the control of a group of intersections instead of a single intersection.

As mentioned above, there were plenty of good techniques for real-time traffic signal control. However, most existing techniques were tested on crossroads rather than modern roundabouts. The modern roundabout has been developed since 1960s and is a common intersection form in many countries nowadays [1][2]. In the literature, roundabouts without signal control have been widely studied and used [22]-[27]. However, without traffic signal to organize the traffic flow, big traffic jams happen at roundabouts in world-wide urban cities every day. Many studies suggested the installation of signal control modules at roundabouts [1][2][28]-[31]. Yang et al. [1][2] and Bai et al. [31] derived mathematical equations to compute the signal cycle length and the green time for each phase. These works provided fixed-time control method for roundabouts.

However, fixed-time signal control methods cannot meet the changing need of the real-world roundabout, which may lower the efficiency of the traffic system. In this paper, we propose to use real-time control method for signalizing the roundabout. First, considering that the FUZZY-TURN and FUZZY-JUMP methods for crossroads have potential to be used on roundabouts, we extend the two methods to the application of roundabouts. As the FUZZY-TURN controller uses well-designed phase sequences only permitting consistent adjacent phases, it is less likely to cause traffic jams even when the traffic is heavy. But the use of preset phase sequences may lower the real-timeness of the controller. On the other side, FUZZY-JUMP is capable of frequently changing its phase sequences and therefore has better real-timeness than FUZZY-TURN. However, the FUZZY-JUMP controller enables inconsistent traffic flows of adjacent phases exist simultaneously and is more likely to cause traffic jams. Moreover, frequently changing sequence of phases may confuse the drivers and lead to traffic accidents.

To overcome the above shortcomings, a mixed fuzzy traffic control mechanism termed FUZZY-MIX is proposed in this paper. The phase set is divided into several phase subsets according to the directions. The FUZZY-MIX has two layers of fuzzy controllers. The outer layer computes the urgency degrees of different phase subsets and determines the next running subset, while the inner layer computes the extension time of the green phase and determines whether to turn to the next phase in the running phase subset. By this way, the FUZZY-MIX technique is a combination of FUZZY-JUMP and FUZZY-TURN methods. The proposed FUZZY-MIX incorporats the advantages of FUZZY-TURN and FUZZY-JUMP with the outer layer to improve real-timeness and the inner layer to reduce the risk of traffic jam and improve traffic safety.

But in a fuzzy controller, the traditional use of membership functions given by human preference cannot guarantee the optimal performance of the controller [13][18][19]. In order to further improve the performance of FUZZY-MIX, this paper develops particle swarm optimization (PSO) to optimize the membership functions adopted in the controller. The PSO algorithm is firstly introduced by Kennedy and Eberhart [32][33] in 1995, and is a relatively new population-based search technique. The algorithm simulates the swarm behaviors of bird flocking or fish schooling and is very easy to implement. Moreover, in recent years the algorithm has been successfully applied on various applications and demonstrated to have high operating efficiency [34]-[36]. PSO is very suitable for solving optimization problems especially for continuous optimization problems. In this paper, the PSO algorithm is applied on optimizing the shapes of the membership functions used in FUZZY-MIX and therefore derives a FUZZY-MIX-OPT controller. In the proposed FUZZY-MIX-OPT controller, the shapes of membership functions are no longer given by human preference, but optimized by the intelligent algorithm inartificially.

In the experiment, a common roundabout with two circulatory lanes and four approaches is simulated, and sixteen traffic conditions including eight steady conditions and eight time-varying conditions are tested. We implement five controllers including the vehicle actuated method termed VA [12], the FUZZY-TURN method [13]-[15], the FUZZY-JUMP method [16]-[17], and the proposed FUZZY-MIX and FUZZY-MIX-OPT methods on the roundabout. Simulation results and comparisons show that the FUZZY-MIX controller outperforms the previous traffic controllers VA, FUZZY-TURN, and FUZZY-JUMP. In addition, by using the PSO algorithm to optimize the membership functions of FUZZY-MIX, the PSO-based FUZZY-MIX-OPT controller further improves the performance of FUZZY-MIX and provides a very effective and efficient traffic signal control technique for the roundabout.

The contribution of this paper is three-fold. First, traffic congestion is a critical problem of modern roundabouts but seldom research has been made on signalizing roundabouts in the literature. This paper develops a real-time signal control method for roundabouts and helps to relieve the traffic burden of modern roundabouts. Second, a novel fuzzy traffic control model with two fuzzy layers incorporating the FUZZY-TURN and FUZZY-JUMP methods is designed. The proposed FUZZY-MIX controller is demonstrated to be very effective to improve the traffic efficiency of the roundabout. Third, we propose the usage of PSO algorithm to optimize the membership functions adopted in FUZZY-MIX and succeed in further improving the performance of the proposed controller.

The rest of this paper is organized as follows. Section II introduces the geometric design and phase composing of the roundabout. Section III describes the implementations of the fuzzy traffic controller for signalizing the roundabout. In Section IV, the PSO algorithm is applied on optimizing the membership functions of the fuzzy traffic controller. In Section V, simulations are conducted and comparisons of five traffic controllers for the roundabout are made. At last, a conclusion is drawn in Section VI.

## II. Roundabout Model

Roundabout is one of the most commonly employed traffic infrastructures in urban cities. According to the environmental issues and traffic conditions, there are various roundabouts with different geometric designs and signal control modules. Roundabout with two circulatory lanes and four approaches is widely used in both real world and academic world [1][2][17]. In this paper, we simulate this kind of roundabout. This section describes the model of the roundabout.

### A. Geometric Design

As shown in Fig. 1, the roundabout has four approaches, each of which has three entrance lanes for each direction. From left to right, the first lane labeled "a" is the left-turn lane occupied by vehicles to turn left, e.g., vehicles from approach 0 to approach 3. The second lane labeled "b" is the go-through lane used for go-through movement, e.g., vehicles from approach 0 to approach 2. The third lane labeled "c" is the right-turn lane which permits vehicles turning right, e.g., vehicles from approach 0 to approach 1. Meanwhile, there are two circulatory lanes for vehicles moving around the roundabout. The inner lane is only used for left-turn traffic flows, while the outer lane is used for both the go-through traffic flows and the left-turn vehicle which is entering or leaving the inner lane. The right-turn vehicle has its unique roadway to pass the roundabout without entering any circulatory lanes.

### B. Traffic Signal

In a signalized roundabout, the stop lines are located on the left-turn and go-through entrance lanes of each approach. Traffic flows behind the stop lines are controlled by traffic signals. Besides, vehicles on the right-turn lane can turn right at any time because they have their unique roadways and are independent with other traffic flows. Thus, for the roundabout, eight traffic flows consisting of the left-turn and go-through flows on each of the four approaches should be taken into consideration. Fig. 2 shows six possible phases which are commonly used for the roundabout. The descriptions of the six phases are listed as follows.

**Phase 0:** the go-through traffic flows on approaches 0 and 2 are activated, whereas the other traffic flows are stopped.

**Phase 1:** only the left-turn traffic flows on approaches 0 and 2 are activated.

**Phase 2:** all the traffic flows on approaches 0 and 2 get green time.

**Phase 3:** green time for approaches 0 and 2 ends. The go-through traffic flows on approaches 1 and 3 are activated.

**Phase 4:** the left-turn flows on approaches 1 and 3 start to move whereas the other flows are stopped.

**Phase 5:** a green time for all the traffic flows on approaches 1 and 3.

The phase set can be divided into two subsets according to the direction. Subset I permits north-south traffic flows (traffic flows from approaches 0 and 2) and stops the west-east traffic flows (traffic flows from approaches 1 and 3), whereas subset II does just the opposite.

As shown in Fig. 3, a phase circle is a cycle of the phase sequences. It is well-designed, only permitting consistent traffic flows in adjacent phase sequences to avoid traffic jam. Traditional traffic signal timing is to determine how long each phase involved in the circle should last. The fixed-time control methods use fixed durations for the phases, whereas the real-time control methods use time-varying durations for the phases. However, using preset phase sequences may lower the real-time response of the system. Thus, instead of using the phase circle, some state-of-the-art real-time signal controllers always jump to the phase which is in the most urgent so as to improve the real-timeness. However, as this mechanism may lead to occurrences of inconsistent traffic flows in adjacent phases (e.g. if the system jumps instantly from phase 0 to phase 5), the system is vulnerable to traffic jam. Employing an all-red time between phase switching so as to clear the vehicles in the roundabout can help to reduce the risk of traffic jam [37][38]. But frequently inserting all-red time induces additional time cost and reduces the traffic capacity of the roundabout during a time period.

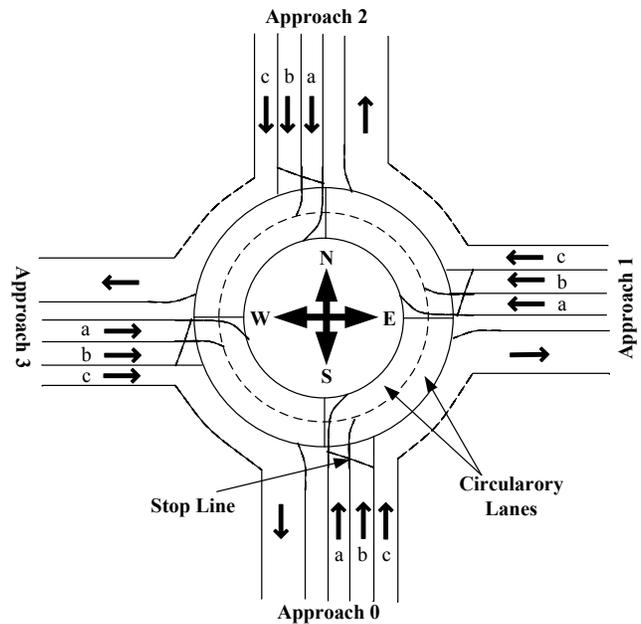

Fig. 1. Illustration of the roundabout.

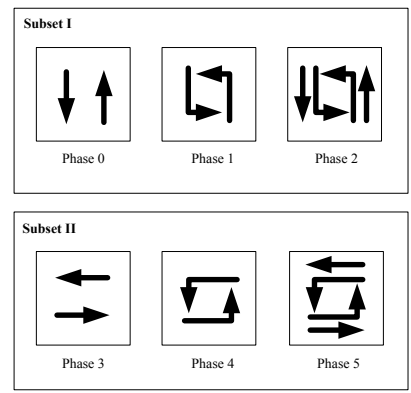

Fig. 2. Phase set of an ordinary four-approach roundabout.

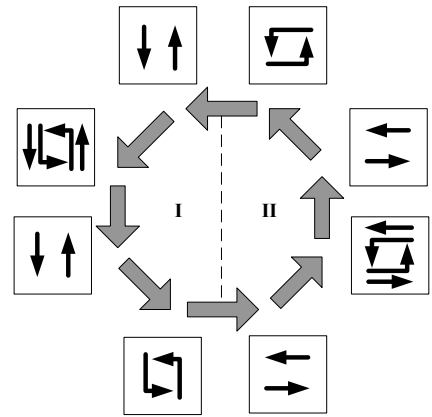

Fig. 3. Phase circle of the four-approach roundabout.

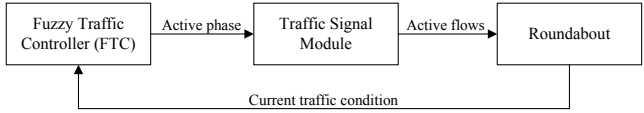

Fig. 4. Illustration of using FTC to signalize roundabout.

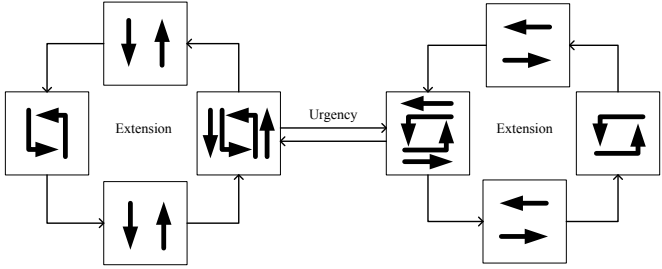

Fig. 5. Phase sequences of FUZZY-MIX controller.

Table I
Notations

| Symbols | Descriptions |
|---|---|
| $ET$ | Extension time |
| $UD$ | Urgency degree |
| $QL$ | Queue length |
| $WT$ | Waiting time |
| $\Phi$ | Threshold value for extending the current phase |
| $\Theta$ | Initial duration for each phase |
| $\Delta$ | Interval time for recomputing $UD$ |
| $\theta$ | All-red time |
| $R_{1\text{-}1}$-$R_{1\text{-}9}$ | Fuzzy rules to infer $ET$ |
| $R_{2\text{-}1}$-$R_{2\text{-}9}$ | Fuzzy rules to infer $UD$ |
| $M$ | Number of particles |
| $N$ | Dimensionality of the search space |
| $\mathbf{X}_i$ | Position vector of particle $i$ |
| $\mathbf{V}_i$ | Velocity vector of particle $i$ |
| $Fit_i$ | Fitness value of particle $i$ |
| $\mathbf{pBest}_i$ | Previous best position of particle $i$ |
| $\mathbf{gBest}$ | Global best position of the whole swarm |
| $\omega$ | Inertia weight |
| $c_1, c_2$ | Acceleration coefficients |
| $G$ | Maximal number of iterations |
| $vehMiss$ | Number of undetected vehicles in the simulation |
| $vehPass$ | Number of passing vehicles in the simulation |
| $vehDelay$ | Average delay time of vehicles in the simulation |
| $w_1, w_2$ | Relative weights used in the evaluation of particles |

## III. FUZZY TRAFFIC CONTROLLER

This section presents the details of using fuzzy traffic controller (FTC) to timing the signal of a roundabout, the illustration of which is shown in Fig. 4. The FTC receives the current traffic condition of the roundabout as input and outputs the active phase to the traffic signal module. According to the selected phase, the traffic signal module allows some traffic flows to move and stops the others behind the stop lines. Then the traffic condition of the roundabout changes, and after a time period the FTC accordingly generates the next active phase. In this paper, a two-layer FTC hybridizing FUZZY-TURN and FUZZY-JUMP is designed. The following of this section first introduces the general method of the proposed FUZZY-MIX and then describes the implementations of the controller. The notations used in this paper are listed in Table I.

### A. General Method

The proposed FUZZY-MIX controller is a mixture of the FUZZY-TURN and FUZZY-JUMP controllers. Therefore, in this part, in order to introduce the general method of FUZZY-MIX, the descriptions of FUZZY-TURN and FUZZY-JUMP are also presented.

**FUZZY-TURN:** by the end of each phase, the FUZZY-TURN generates an extension time ($ET$) of the current phase according to the traffic condition of the roundabout. If $ET$ is larger than a predefined threshold value ($\Phi$ time units), the green phase lasts for another $ET$ period of time. Otherwise, the current green phase is terminated and the next phase in the phase circle (shown in Fig. 3) is activated and allocated with an initial duration ($\Theta$ time units).

**FUZZY-JUMP:** at set intervals ($\Delta$ time units), the urgency degree ($UD$) of each phase is computed. Then the signal module jumps to the phase with largest $UD$. Obviously, if the current green phase has the largest $UD$, it lasts for another $\Delta$ period of time; otherwise it is replaced by another phase. What should be pointed out is that if the adjacent phases are inconsistent, an all-red time ($\theta$ seconds) should be inserted between the two phases to avoid traffic jam.

**FUZZY-MIX:** the phase set is divided into two subsets as presented in Section II-B. The controller is composed of two layers. The outer layer is a coarser-grained form of the FUZZY-JUMP controller. It computes the $UD$ of the two phase subsets by the end of each phase. If the current running subset has a larger *UD* than the other subset, the inner layer is activated. The inner layer is similar to the FUZZY-TURN controller, which computes the *ET* of the current green phase. If *ET* is larger than $\Phi$, the phase is kept unchanged; otherwise the next phase in the current subset is activated and initialized with $\Theta$ time units. The phase sequences of the two subsets are shown in Fig. 5. On the other hand, if the current running subset has a smaller *UD* than the other subset, the system will execute an all-red time and then jumps to the entrance phase of the other subset. As shown in Fig. 5, the entrance phases of the two subsets are the phases with maximum active flows, i.e., phase 1 in subset I and phase 4 in subset II.

### B. Implementations of FUZZY-MIX

A fuzzy logic controller [39][40] consists of a fuzzification module, a fuzzy rule base, an inference engine, and a defuzzification module. The fuzzification module performs membership functions that convert the crisp input values into linguistic values with fuzzy membership grades. The inference engine receives the fuzzification results and employs the rules in the fuzzy rule base to infer fuzzy (linguistic) outputs. It is capable of simulating human decision-making and is the kernel of the fuzzy logic controller. At last, the defuzzification module converts the fuzzy outputs into a crisp value.

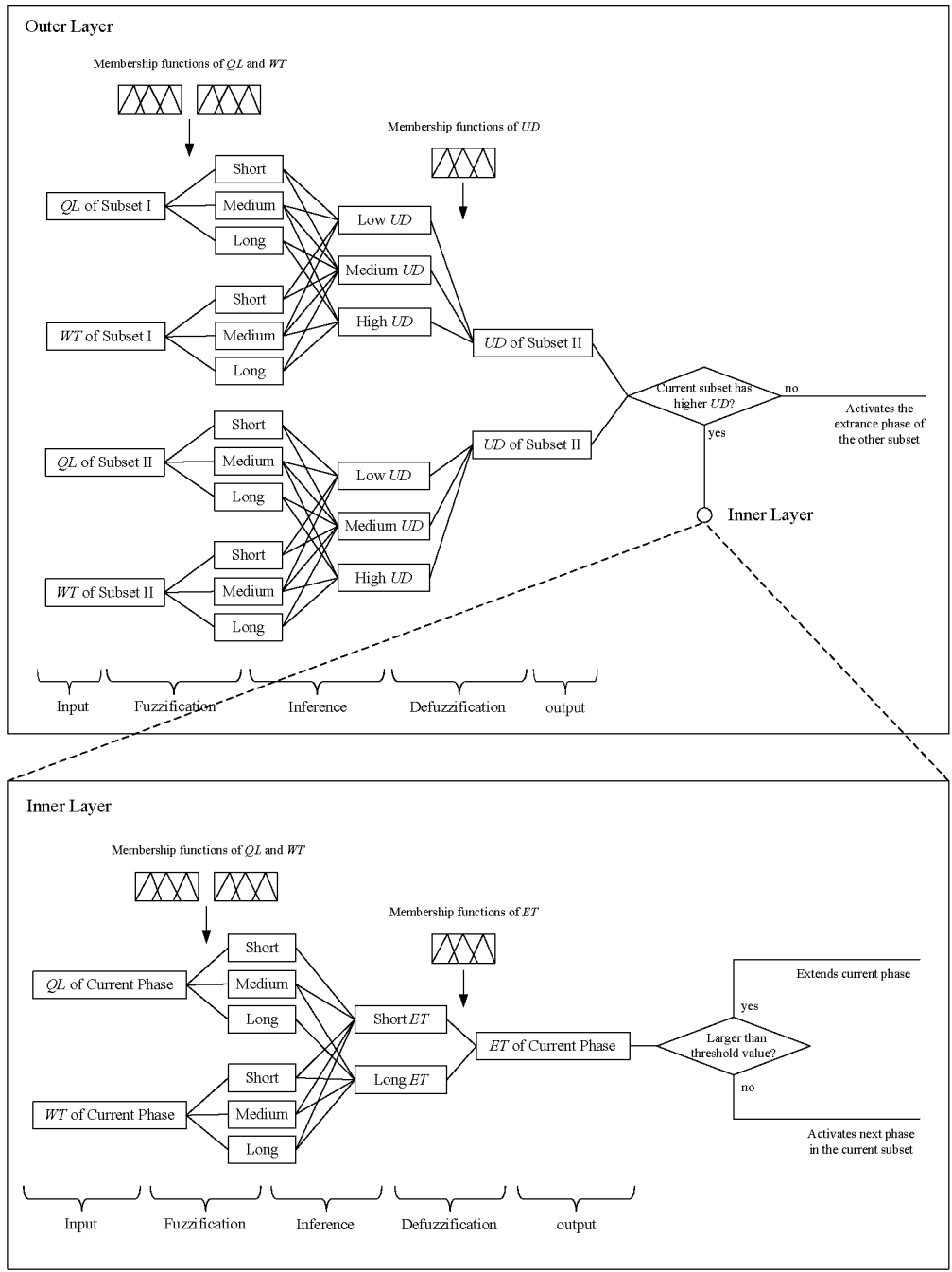

Fig. 6. Schematic diagram of FUZZY-MIX controller.

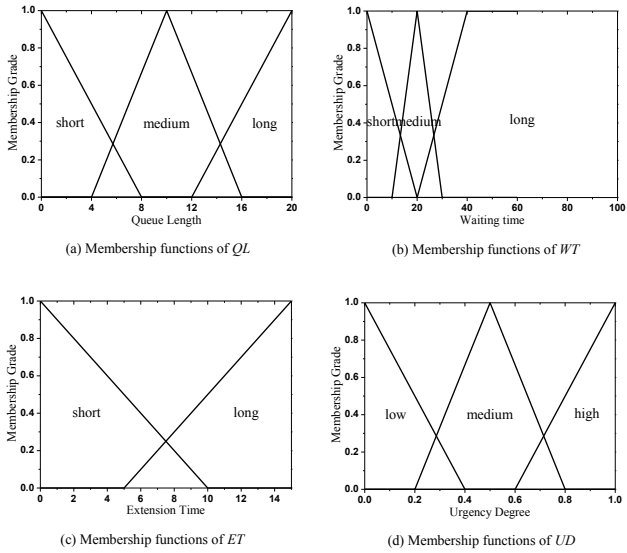

Fig. 7. Membership functions of fuzzy variables.

Table II
Fuzzy Rule Base

| Rule | Inputs | | Output 1 | Output 2 |
|---|---|---|---|---|
| | QL | WT | ET | UD |
| $R_{1-1}/R_{2-1}$ | short | short | short | low |
| $R_{1-2}/R_{2-2}$ | short | medium | short | low |
| $R_{1-3}/R_{2-3}$ | short | long | short | medium |
| $R_{1-4}/R_{2-4}$ | medium | short | short | low |
| $R_{1-5}/R_{2-5}$ | medium | medium | long | medium |
| $R_{1-6}/R_{2-6}$ | medium | long | long | high |
| $R_{1-7}/R_{2-7}$ | long | short | long | medium |
| $R_{1-8}/R_{2-8}$ | long | medium | long | high |
| $R_{1-9}/R_{2-9}$ | long | long | long | high |

The schematic diagram of the proposed FUZZY-MIX system is shown in Fig. 6. It can be observed that the system is composed of two layers of fuzzy logic controllers. The outer layer computes the urgency degrees of the two phase subsets. Only when the current phase subset is in more urgent than the other one, the inner layer is activated to compute the extension time of the current phase. Otherwise, the controller activates the entrance phase of the other phase subset. Details of the implementations are described as follows.

*1) Variables and Membership Functions*

Fuzzy variables used in the controller are: queue length (*QL*) and waiting time (*WT*) as inputs to reflect the current traffic condition, and *ET* and *UD* as outputs to control the traffic signal. *QL* of a phase is the average number of vehicles detected behind the stop lines which have green signal in the phase, and *WT* is the average duration of these vehicles since they have arrived the roundabout. Moreover, *QL* and *WT* of a phase subset are represented by those of its entrance phase. Besides, *ET* and *UD* are described in the above Part A of this Section.

Membership functions of the four input/output variables are shown in Fig. 7, in which the *x*-axis is the quantity of a variable while the *y*-axis is the corresponding membership grade. It can be observed in Fig. 7 (a) that *QL* has three membership functions including *short_QL*(*x*), *medium_QL*(*x*), and *long_QL*(*x*) as defined in (1)-(3).

$$short\_QL(x) = \begin{cases} (8-x)/8, & \text{if } x < 8 \\ 0, & \text{if } x \geq 8 \end{cases} \quad (1)$$

$$medium\_QL(x) = \begin{cases} 0, & \text{if } x \leq 4 \text{ or } x \geq 16 \\ (x-4)/6, & \text{if } 4 < x < 10 \\ (16-x)/6, & \text{if } 10 \leq x < 16 \end{cases} \quad (2)$$

$$high\_QL(x) = \begin{cases} 0, & \text{if } x \leq 12 \\ (x-12)/8, & \text{if } x > 12 \end{cases} \quad (3)$$

In the same way, *WT* has three membership functions including *short_WT*(*x*), *medium_WT*(*x*), and *long_WT*(*x*); *ET* has two membership functions including *short_ET*(*x*) and *long_ET*(*x*); and *UD* has three membership functions including *low_UD*(*x*), *medium_UD*(*x*), and *high_UD*(*x*). These membership functions are illustrated in Fig.7 (b), (c), and (d) respectively. In the fuzzification module, the membership functions of input variables *QL* and *WT* are performed to fuzzify the crisp inputs in order to obtain the corresponding membership grades. On the other hand, the membership functions of output variables *ET* and *UD* are used in the defuzzification module, which will be described later.

*2) Fuzzy Rule Base and Fuzzy Inference*

Table II shows the fuzzy rule base. It is composed of eighteen rules including nine rules ($R_{1-1}$-$R_{1-9}$) to infer *ET* and nine rules ($R_{2-1}$-$R_{2-9}$) to infer *UD*. For example, $R_{1-1}$ means that `if` *QL* `is` short and *WT* `is` short `then` *ET* `is` short; $R_{2-1}$ means that `if` *QL* `is` short and *WT* `is` short `then` *UD* `is` low; $R_{1-7}$ stands for that `if` *QL* `is` long and *WT* `is` short `then` *ET* `is` long; $R_{2-7}$ stands for that `if` *QL* `is` long and *WT* `is` short `then` *UD* `is` medium.

Then, the fuzzy inference is applied to combine these `"if-then"` rules into a mapping from fuzzy input set to fuzzy output set. In the inference engine of FUZZY-MIX, rule matching is conducted by using fuzzy intersection (the min operator), while rule merging is done by employing fuzzy union (the max operator). For example, suppose that the membership grade of "*QL* `is` short" is $u_1$ and the membership grade of "*WT* `is` short" is $u_2$, the inference strength of $R_{1-1}$ is computed by $r_{11} = \min\{u_1, u_2\}$. Moreover, as "*ET* `is` short" is the consequent of $R_{1-1}$, $R_{1-2}$, $R_{1-3}$, and $R_{1-4}$, the membership grade of "*ET* `is` short" is computed by $o_1 = \max\{r_{11}, r_{12}, r_{13}, r_{14}\}$.

*3) Defuzzification*

Finally, the defuzzification module is executed. It is necessary because the traffic system needs crisp signal timings. A height defuzzification method [41] is adopted, in which the resultant output is sensitive to all the corresponding rules executed. At first, the centroid of each membership functions is computed. Then the final crisp output is the average of these centroids, weighted by the membership grades (heights). The output of the fuzzyfication is defined as

$$O = \frac{\sum_{i=1}^{k} o_i \times c_i}{\sum_{i=1}^{k} o_i} \quad (4)$$

where $k$ is the number of membership functions of the output variable; $o_i$ is the membership grade of the $i$th fuzzy output, $c_i$ is the centroid of the $i$th membership function. For example, suppose that the membership grades of "ET is short" and "ET is long" are 0.408 and 0.083 respectively, as the centroid of $short\_ET(x)$ and $long\_ET(x)$ are 2.5 and 12.5 respectively, the final crisp output of ET is computed as $O_{ET} = (0.408 \times 2.5 + 0.083 \times 12.5)/(0.408 + 0.083) = 4.19$.

## IV. PSO-BASED FUZZY TRAFFIC CONTROLLER

In industrial applications of fuzzy logic controller, the shapes of membership functions are always chosen by human arbitrarily. It is based on engineers' experience, and cannot guarantee to provide optimal control for the corresponding system. In this paper, to further improve the performance of the FUZZY-MIX controller, a PSO algorithm is adopted to optimize the shapes of the membership functions. The FUZZY-MIX controller with optimal membership functions trained by PSO is then termed FUZZY-MIX-OPT. This Section describes the implantations of applying PSO algorithm for optimizing the membership functions of FUZZY-MIX-OPT.

### A. Methodology

PSO algorithm stimulates the foraging behavior of birds or fishes, in which a group of particles are randomly scattered in the problem space and search for the optimal point simultaneously. In PSO algorithm, a particle swarm composed of $M$ particles is maintained. Each particle is associated with a position vector $X_i = [x_i^1, x_i^2, ..., x_i^N]$, a velocity vector $V_i = [v_i^1, v_i^2, ..., v_i^N]$, and a fitness value $Fit_i$ (where $N$ is the dimensionality of the search space and $i = 1, 2, ..., M$). The position vector $X_i$ stands for a candidate solution of the optimization problem, and is evaluated to obtain the fitness value of a particle. Each particle maintains a previous best position vector $pBest_i = [pBest_i^1, pBest_i^2, ..., pBest_i^N]$ as its search experience. Meanwhile, the global best position found by the whole swarm during the search process is recorded as $gBest = [gBest^1, gBest^2, ..., gBest^N]$. Then, particle $i$ adjusts its flying velocity based on $pBest_i$ (self-cognitive) and $gBest$ (social-cognitive), and accordingly updates its position vector. By this way, the particles tend to fly towards better and better domain. Eventually, the particle swarm is likely to converge to the optimal position of the problem space. PSO is capable of exploring and exploiting some large, complex, and initially unknown problem spaces, which traditional algorithms (enumerative, heuristic, etc.) can hardly deal with. The PSO algorithm is very suitable to optimize the membership functions of the fuzzy traffic controller.

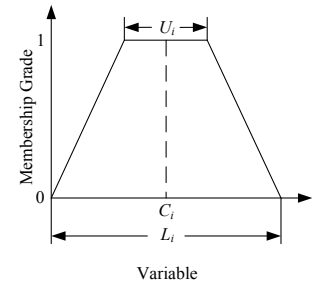

Fig. 8. Representation of a trapezoidal function.

### B. Representation

In order to use PSO algorithm for optimizing the membership functions in FUZZY-MIX-OPT, we should define the representation of particle's position and the search space. Each particle's position should indicate a candidate solution of the problem, i.e., the shapes of all the membership functions adopted in FUZZY-MIX-OPT.

As described in Section III, there are totally eleven membership functions in FUZZY-MIX including three for QL, three for WT, two for ET, and three for UD. All the membership functions are triangular or trapezoidal functions, which are commonly used in fuzzy logic controllers. Moreover, a triangular function can be regarded as a special trapezoidal function (the length of upper line is 0). Therefore, only trapezoidal function is concerned in the optimization.

As shown in Fig. 8, each membership function in FUZZY-MIX-OPT can be represented by a triple $<U_i, L_i, C_i>$, where $U_i$ and $L_i$ stand for the length of upper line and lower line respectively, and $C_i$ is the coordinate of the middle point of the parallel sides. It should be noticed that each membership function is regarded as an isosceles trapezoid in order to reduce the search space of the PSO algorithm. In the experiment, we will find that this simplification does not lower the performance of the proposed FUZZY-MIX-OPT controller. Moreover, in each trapezoidal function there exists a constraint that $U_i$ should be no larger than $L_i$. We can eliminate this constraint by introducing a variable $D_i = L_i - U_i$ and representing each trapezoidal function as $<U_i, D_i, C_i>$. In the representation of PSO algorithm, each candidate solution (particle's position) is in a form of $[U_1 D_1 C_1 U_2 D_2 C_2 \ldots C_{11}]$, which indicates the shapes of the 11 membership functions in the FTC. Consequently, the dimensionality of the search space of the PSO algorithm is 33. In addition, the ranges of all the variables are presented as follows,

$0 \leq U_i \leq 10,\ 0 \leq D_i \leq 10,\ 0 \leq C_i \leq 20\ (i = 1, 2, 3)$,

$$0 \leq U_i \leq 50, \ 0 \leq D_i \leq 50, \ 0 \leq C_i \leq 100 \ (i=4,5,6),$$
$$0 \leq U_i \leq 7.5, \ 0 \leq D_i \leq 7.5, \ 0 \leq C_i \leq 15 \ (i=7,8),$$
$$0 \leq U_i \leq 0.5, \ 0 \leq D_i \leq 0.5, \ 0 \leq C_i \leq 1 \ (i=9,10,11).$$

*C. Evaluation*

In the algorithm, in order to find which candidate solution showing good performance, the positions of all the particles should be evaluated. In the literature, the fitness value of each particle is commonly evaluated by a mathematical function (called fitness function) defined according to the objective of the problem. However, it is hard to define a mathematical fitness function for a traffic signal control system. Therefore, in this paper we perform simulations and evaluate each particle's position by the performance measure during the simulation.

In the algorithm, $Fit_i$ of particle $i$ is determined by

$$Fit_i = w_1 \times (vehMiss/vehPass) + w_2 \times vehDelay \quad (5)$$

where $vehMiss/vehPass$ is the ratio of the number of undetected vehicles ($vehMiss$) to the number of passing vehicles ($vehPass$) during the simulation; $vehDelay$ stands for the average delay time of vehicles; $w_1$ and $w_2$ are the corresponding weights for $vehMiss/vehPass$ and $vehDelay$. Details of the simulation will be described in Section V.

*D. Overall Procedure*

The procedure of the PSO algorithm for optimizing membership functions of FUZZY-MIX-OPT is very simple, which is shown in Fig. 9. In the initialization, the position and velocity vectors of the $M$ particles are randomly generated. The fitness values of the particles are evaluated according to (5). The $pBest_i$ of particle $i$ is set with its initial position vector $X_i$, and then the $gBest$ is set with the currently best $pBest_i$.

In each iteration of the algorithm, the particles in the swarm interact with each other and collaborate to search the problem space. It should be noticed that a particle by itself can hardly solve any problem, and that the evolvement happens just when the particles interact. The update of position and velocity for particle $i$ is defined as

$$V_i = \omega V_i + c_1 rand_1 \otimes (pBest_i - X_i) + c_2 rand_2 \otimes (gBest - X_i), \quad (6)$$

$$X_i = X_i + V_i, \quad (7)$$

where $\omega$ is the inertia weight; $c_1$ and $c_2$ are acceleration coefficients which determine the relative weights of self-cognitive and social-influence; $rand_1$ and $rand_2$ are two random vectors uniformly distributed over $[0, 1]^N$; $\otimes$ is component-wise multiplication. In equation (6), $\omega V_i$ can be regarded as the inertia of particle $i$'s velocity, while $c_1 rand_1 \otimes (pBest_i - X_i) + c_2 rand_2 \otimes (gBest - X_i)$ can be interpreted as the external force to pull the particle towards better position in the problem space. It has been proved in the literature [42] that $\omega$ plays a role of balancing the exploration and exploitation of PSO algorithm. A good performance can be gained by using a large $\omega$ (e.g., 0.9) at beginning to explore the search space, and gradually reducing $\omega$ to a lower value (e.g., 0,4) to refine the solution. After updating the velocity of particle $i$, the position of the particle is updated according to (7).

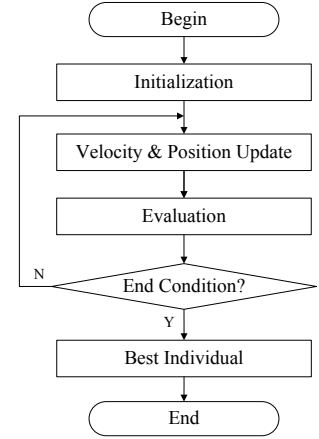

Fig. 9. Flowchart of PSO algorithm.

Then, the positions of all the particles are evaluated. If the new position of particle $i$ is better than its previous best position, $pBest_i$ is replaced by $X_i$. Furthermore, if a new global best-so-far position is discovered, the $gBest$ is accordingly updated. After a few iterations, the particle swarm converges, and an optimal or near-optimal solution of the problem is obtained. The PSO algorithm is always defined with a maximal number of iterations $G$, by which the algorithm will be terminated.

V. SIMULATION RESULTS AND COMPARISONS

In the experiment, the performance of different controllers is compared by simulations conducted on the roundabout with two circulatory lanes and four approaches. In order to comprehensively investigate the performance of these controllers for signalizing the roundabout, we generate sixteen traffic conditions for the roundabout, each of which has its own characteristics. Then, five real-time traffic controllers including the vehicle actuated method (VA) [12], the FUZZY-TURN controller, the FUZZY-JUMP controller, the FUZZY-MIX controller, and the FUZZY-MIX-OPT controller are applied. In this Section, we first describe the simulation environment and parameters, and then present the simulation results and comparisons.

*A. Simulation Environment and Parameters*

The simulation program is developed with Visual C++, run on a machine with Intel Pentium Dual CPU, 1.99 GHz/500 MB of RAM. The geometric design and phase composing of the roundabout are shown in Fig. 1 and Fig. 2 respectively and described in Section II. The arrivals of vehicles on different lanes are independent and subject to Poisson distribution [19]. We perform the simulations on sixteen different traffic conditions consisting of eight steady conditions and eight time-varying conditions. The Poisson arrival rates of vehicles on each entrance lanes of these sixteen conditions are shown in Table III, where 0-L, 1-L, 2-L, and 3-L stand for the left-turn entrance lanes of approaches 0, 1, 2, and 3 respectively; 0-S, 1-S, 2-S, and 3-S represent the go-straight entrance lanes of approaches 0, 1, 2, and 3 respectively; a time unit is equal to 0.5 seconds. For example, if condition C1 is tested, the probability of a vehicle being generated on the left-turn lane of approach 0 per time unit is 0.102. It means that the expectation of the number of vehicles generated on this entrance lane per hour is $0.102 \times 2 \times 3600 = 734$. Besides, when a time-varying condition is applied, the Poisson arrival rate of the vehicle on each entrance lane will start at a beginning value and smoothly increase to an ending value over time.

The differences among the sixteen traffic conditions are presented as follows.

1. The flow rates of C1-C8 are steady whereas those of C9-C16 are time-varying.
2. In C1, C3, …, C15, the differences between the north-south flow rates (the arrival rates of vehicles from approaches 0 and 2) and the west-east flow rates (the arrival rates of vehicles from approaches 1 and 3) are inconspicuous. In contrast, in C2, C4, …, C16, those differences are conspicuous.
3. In C1, C2, C5, C6, C9, C10, C13, and C14, the left-turn flow rate and go-straight flow rate of a same approach are similar. In contrast, the left-turn and go-straight flow rates of a same approach differ a lot in the other conditions.
4. Among the steady traffic conditions, C1-C4 are light conditions the vehicle arrival rates of which are relatively low, whereas C5-C8 are heavy conditions which have much higher vehicle arrival rates than C1-C4.
5. Among the time-varying conditions, the range of the flow rate in C13-C16 is larger than that in C9-C12.

It is assumed that the detectors are installed at a certain distance to the roundabout, and the maximal number of detectable queuing vehicles of each entrance lane is equal to 20. If the queue length of an entrance lane exceeds 20, the newly arrived vehicle on this lane will be regarded as an undetected vehicle. We always expect to reduce the length of waiting-vehicle queue and the number of undetected vehicles (*vehMiss*). Moreover, the undetected vehicles lead to occurrence of detecting error which would do harm to the performance of the fuzzy controller. Therefore, when using PSO algorithm to optimize the membership functions of FUZZY-MIX-OPT, minimizing the *vehMiss* should be concerned. Due to the random factors in the simulation, minimizing the ratio of *vehMiss* to *vehPass* is used as the first objective of PSO, which is shown in equation (5). Moreover, as a common performance index of traffic signal control, the average delay time of vehicles is used as the second objective of PSO. In the experiment, we set $w_1$ and $w_2$ of equation (5) as $w_1=1$ and $w_2=10^{-8}$ so as to make minimizing the *vehMiss*/*vehPass* in preference to minimizing the *vehDelay*.

Table III
Sixteen Traffic Conditions

| Condition | | 0-L | 1-L | 2-L | 3-L | 0-S | 1-S | 2-S | 3-S |
|---|---|---|---|---|---|---|---|---|---|
| Steady | C1 | 0.102 | 0.097 | 0.092 | 0.123 | 0.118 | 0.108 | 0.108 | 0.125 |
| | C2 | 0.156 | 0.099 | 0.143 | 0.102 | 0.176 | 0.111 | 0.180 | 0.108 |
| | C3 | 0.067 | 0.074 | 0.068 | 0.068 | 0.178 | 0.149 | 0.158 | 0.169 |
| | C4 | 0.057 | 0.111 | 0.069 | 0.121 | 0.154 | 0.199 | 0.142 | 0.212 |
| | C5 | 0.178 | 0.165 | 0.189 | 0.215 | 0.199 | 0.212 | 0.203 | 0.222 |
| | C6 | 0.181 | 0.121 | 0.243 | 0.132 | 0.209 | 0.131 | 0.255 | 0.143 |
| | C7 | 0.132 | 0.114 | 0.117 | 0.108 | 0.251 | 0.232 | 0.223 | 0.209 |
| | C8 | 0.101 | 0.188 | 0.098 | 0.200 | 0.198 | 0.287 | 0.216 | 0.320 |
| Time-varying | C9 | 0.089 ↓ 0.178 | 0.102 ↓ 0.177 | 0.103 ↓ 0.190 | 0.077 ↓ 0.168 | 0.112 ↓ 0.198 | 0.108 ↓ 0.189 | 0.131 ↓ 0.179 | 0.105 ↓ 0.170 |
| | C10 | 0.057 ↓ 0.224 | 0.066 ↓ 0.232 | 0.072 ↓ 0.189 | 0.072 ↓ 0.250 | 0.080 ↓ 0.200 | 0.066 ↓ 0.240 | 0.073 ↓ 0.232 | 0.081 ↓ 0.255 |
| | C11 | 0.156 ↓ 0.223 | 0.078 ↓ 0.182 | 0.123 ↓ 0.189 | 0.050 ↓ 0.132 | 0.158 ↓ 0.240 | 0.089 ↓ 0.194 | 0.149 ↓ 0.278 | 0.101 ↓ 0.201 |
| | C12 | 0.101 ↓ 0.242 | 0.058 ↓ 0.179 | 0.097 ↓ 0.199 | 0.034 ↓ 0.155 | 0.125 ↓ 0.252 | 0.077 ↓ 0.177 | 0.102 ↓ 0.280 | 0.073 ↓ 0.156 |
| | C13 | 0.101 ↓ 0.158 | 0.097 ↓ 0.135 | 0.108 ↓ 0.138 | 0.099 ↓ 0.160 | 0.159 ↓ 0.277 | 0.188 ↓ 0.256 | 0.180 ↓ 0.280 | 0.176 ↓ 0.310 |
| | C14 | 0.057 ↓ 0.158 | 0.077 ↓ 0.170 | 0.079 ↓ 0.120 | 0.100 ↓ 0.155 | 0.140 ↓ 0.300 | 0.161 ↓ 0.341 | 0.151 ↓ 0.400 | 0.130 ↓ 0.299 |
| | C15 | 0.055 ↓ 0.101 | 0.089 ↓ 0.151 | 0.048 ↓ 0.078 | 0.100 ↓ 0.162 | 0.121 ↓ 0.178 | 0.189 ↓ 0.298 | 0.148 ↓ 0.210 | 0.210 ↓ 0.250 |
| | C16 | 0.055 ↓ 0.158 | 0.102 ↓ 0.189 | 0.077 ↓ 0.176 | 0.134 ↓ 0.188 | 0.160 ↓ 0.298 | 0.205 ↓ 0.315 | 0.144 ↓ 0.298 | 0.245 ↓ 0.341 |

Besides, in the PSO algorithm, the population size $M$ is set as 20; the dimensionality $N$ is 33; the maximal number of iterations $G$ is set as 1,000; $\omega$ is initialized as 0.9 and linearly decreased to 0.4 during the search process; acceleration coefficients are set as that $c_1 = c_2 = 2.0$.

Each simulation lasts 100,000 time units (about 14 h). In the controllers, the initial duration $\Theta$ for each phase is set as 10, the threshold value $\Phi$ for extending the green phase is set as 5, the interval time $\Delta$ for recomputing $UD$ is set as 10, and the length $\theta$ of all-red time is set as 5. The membership functions of FUZZY-TURN, FUZZY-JUMP, and FUZZY-MIX are shown in Fig. 7 which are set by human perception, whereas those of FUZZY-MIX-OPT are set according to the optimization results of the PSO algorithm.

*B. Results Comparisons on Number of Undetected Vehicles*

As an index of performance, the number of undetected vehicles during the simulation is shown in Table IV. Vehicle actuated method has good performance in light and steady conditions C1-C4, but is inferior to the other four controllers in controlling conditions C5-C16. This is because the VA method adjusts the signal just according to whether there is at least one vehicle has been detected in the last time period. It employs little information of the roundabout and is thus limited in controlling heavy and complex traffic conditions. FUZZY-TURN and FUZZY-JUMP do better than VA. In addition, in the comparison of these two controllers, it can be observed that in controlling all the even numbered conditions, FUZZY-JUMP obtains less *vehMiss* than that obtained by FUZZY-TURN. As described in Part A of this Section, in even numbered conditions the differences between the north-south flow rates and the west-east flow rates are conspicuous. The FUZZY-JUMP always changes its green phase to the currently most urgent phase. The controller is much easier to switch the phase from one direction to the other direction than the FUZZY-TURN (which complies with predefined phase circle). Therefore, FUZZY-JUMP is more readily adapt to changing needs and circumstances and outperforms the FUZZY-TURN in controlling even numbered conditions. However, for some odd numbered conditions, the performance of FUZZY-TURN is better than FUZZY-JUMP. This may be because the FUZZY-TURN uses consistent adjacent phase sequences, and saves the cost of all-red time. When the urgency of phases between different directions is similar, the better real-timeness of FUZZY-JUMP is limited and the advantages of FUZZY-TURN emerge. But in general, FUZZY-JUMP obtains less *vehMiss* than FUZZY-TURN in most cases.

As a mixture of FUZZY-TURN and FUZZY-JUMP, the proposed FUZZY-MIX controller outperforms FUZZY-TURN and FUZZY-JUMP for all the tested conditions. The reasons are summarized as follows. First, the inner layer of FUZZY-MIX uses consistent phase sequences, which helps to smooth the traffic flows and reduce congestion. Second, by the use of the outer layer, FUZZY-MIX is capable of breaking the phase sequences if the other direction of the roundabout is in more urgent than the current direction. By this way, the proposed controller can immediately respond to the current need of the roundabout. Third, compared to FUZZY-JUMP which considers the urgent degrees among the phases, the FUZZY-MIX concerns the urgent degrees among the phase subsets. The coarse-grained FUZZY-MIX controller does not change the phase sequences as frequent as the FUZZY-JUMP controller, and therefore induces less all-red time than the FUZZY-JUMP. The decrease in all-red time helps to improve the traffic efficiency of the roundabout. Last, when switching the phase from one phase subset to the other, the entrance phase of the subset is set as the phase which has maximum active flows. As the all-red time (before the entrance phase starts up) clears the vehicles in the circulatory lanes of the roundabout, the vehicles of the urgent direction can occupy the circulatory lanes as soon as possible. This mechanism helps to maximally relieve the traffic burden of the urgent direction.

In addition, by using PSO algorithm to optimize the membership functions adopted in FUZZY-MIX, the FUZZY-MIX-OPT controller further improves the performance of the proposed controller. These results also demonstrate that the traditional use of artificial membership functions does not provide the optimal performance of fuzzy systems. Therefore, using intelligent algorithms to optimize the membership functions of fuzzy logic controllers is promising. As shown in Table IV, the FUZZY-MIX-OPT controller has the best performance among the five compared controllers. In controlling 14 out of the 16 tested conditions, the number of undetected vehicles obtained by FUZZY-MIX-OPT is 0. Only in controlling C14 and C16, the undetected vehicles still exist. This may due to that, in the two conditions, the flow rates increase to very big values, and the traffic volume approaches the capacity of the roundabout. In such cases, the PSO algorithm has little room to make further improvement for reducing the undetected vehicles.

Table IV
Number of Undetected Vehicles Obtained by the Five Controllers

| Condition | VA | FUZZY-TURN | FUZZY-JUMP | FUZZY-MIX | FUZZY-MIX-OPT |
|---|---|---|---|---|---|
| C1 | 0 | 0 | 0 | 0 | 0 |
| C2 | 0 | 4 | 1 | 0 | 0 |
| C3 | 0 | 0 | 0 | 0 | 0 |
| C4 | 7 | 10 | 1 | 0 | 0 |
| C5 | 7988 | 2361 | 825 | 0 | 0 |
| C6 | 1248 | 1345 | 751 | 19 | 0 |
| C7 | 1168 | 239 | 104 | 0 | 0 |
| C8 | 8343 | 5392 | 4603 | 69 | 0 |
| C9 | 161 | 5 | 24 | 0 | 0 |
| C10 | 3474 | 1250 | 858 | 8 | 0 |
| C11 | 1732 | 752 | 460 | 4 | 0 |
| C12 | 1242 | 509 | 409 | 13 | 0 |
| C13 | 4242 | 1082 | 856 | 0 | 0 |
| C14 | 8908 | 4053 | 4038 | 715 | 77 |
| C15 | 404 | 161 | 197 | 4 | 0 |
| C16 | 10708 | 4530 | 4508 | 19 | 3 |

Table V
Average Vehicle Delay (Seconds) Obtained by the Five Controllers

| Condition | VA | FUZZY-TURN | FUZZY-JUMP | FUZZY-MIX | FUZZY-MIX-OPT |
|---|---|---|---|---|---|
| C1 | 39.980 | 39.887 | 45.626 | 32.409 | 22.955 |
| C2 | 44.018 | 43.091 | 45.237 | 31.144 | 24.629 |
| C3 | 41.761 | 40.152 | 41.405 | 32.742 | 23.276 |
| C4 | 45.175 | 42.478 | 42.030 | 32.136 | 23.822 |
| C5 | 80.371 | 73.933 | 62.440 | 31.626 | 27.504 |
| C6 | 61.905 | 58.019 | 54.004 | 31.092 | 26.861 |
| C7 | 61.146 | 51.012 | 50.374 | 30.843 | 26.484 |
| C8 | 74.525 | 71.102 | 70.449 | 38.328 | 28.459 |
| C9 | 48.827 | 44.331 | 47.782 | 31.438 | 24.864 |
| C10 | 60.074 | 56.509 | 56.411 | 32.821 | 25.947 |
| C11 | 58.689 | 54.010 | 52.467 | 30.875 | 26.029 |
| C12 | 54.914 | 50.694 | 51.065 | 31.405 | 25.318 |
| C13 | 64.158 | 54.752 | 55.417 | 31.347 | 26.540 |
| C14 | 66.712 | 58.557 | 63.787 | 35.053 | 29.055 |
| C15 | 52.196 | 46.764 | 46.027 | 32.552 | 25.069 |
| C16 | 72.551 | 65.411 | 68.034 | 33.525 | 28.694 |
| **Mean** | **57.938** | **53.169** | **53.285** | **32.459** | **25.969** |

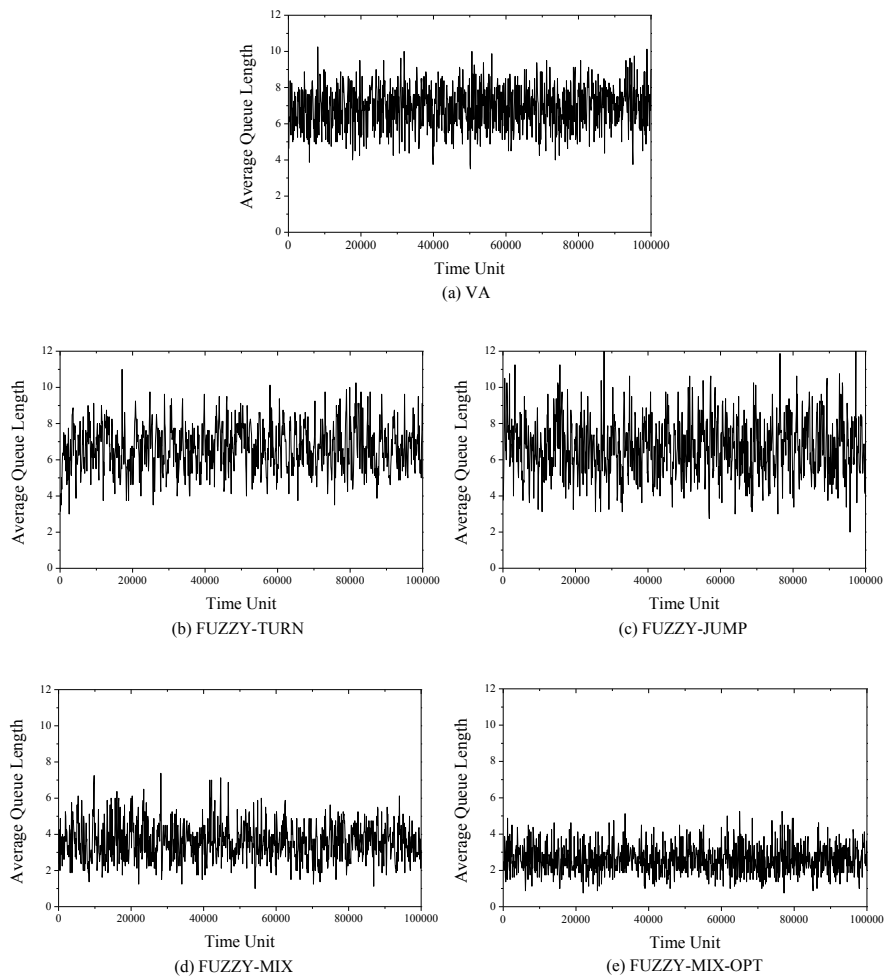

Fig. 10. Average queue length during the simulation of condition C8.

*C. Results Comparisons on Average Vehicle Delay*

In order to further compare the performance of the controllers, the average delay time of vehicles is computed, which is presented in Table V. The last column of Table V shows the mean of *vehDelay* among the sixteen conditions. From Table V, it can be observed that VA has the largest vehicle delay, followed by FUZZY-JUMP and FUZZY-TURN. The proposed FUZZY-MIX and FUZZY-MIX-OPT achieve the smallest delay. The mean of *vehDelay* obtained by FUZZY-MIX-OPT is 25.969 seconds, which is about half of the value obtained by VA (57.938 seconds), FUZZY-TURN (53.169 seconds), and FUZZY-JUMP (53.285 seconds). Moreover, compared with FUZZY-MIX with artificial membership functions (32.459 seconds), the vehicle delay time of FUZZY-MIX-OPT decreases 20%. Therefore, the use of optimal membership functions trained by the PSO algorithm is very effective to improve the traffic efficiency of the roundabout.

In addition, the average length of the eight waiting-vehicle queues of the roundabout during the simulation is shown in Fig. 10 and Fig. 11, where steady condition C8 and time-varying condition C16 are taken as examples. It can be observed that the FUZZY-MIX-OPT controller has the shortest average queue length during the simulation, which further demonstrates the effectiveness and efficiency of the proposed optimal traffic signal controller.

VI. CONCLUSION

In the literature, most of signal control methods are developed for crossroads rather than modern roundabouts. However, without a good signal control, the traffic efficiency of roundabouts in world-wide urban cities is always unsatisfactory. This paper aims to develop real-time traffic signal control method for modern roundabouts. At first, previous FUZZY-TURN and FUZZY-JUMP controllers for crossroads are introduced to control the signalized roundabout. Then, a novel FUZZY-MIX controller which has two fuzzy layers hybridizing FUZZY-TURN and FUZZY-JUMP is designed. Furthermore, the PSO algorithm is applied to optimize the membership functions of FUZZY-MIX. Therefore, a PSO-based

Technical Report – SYSU – 201103

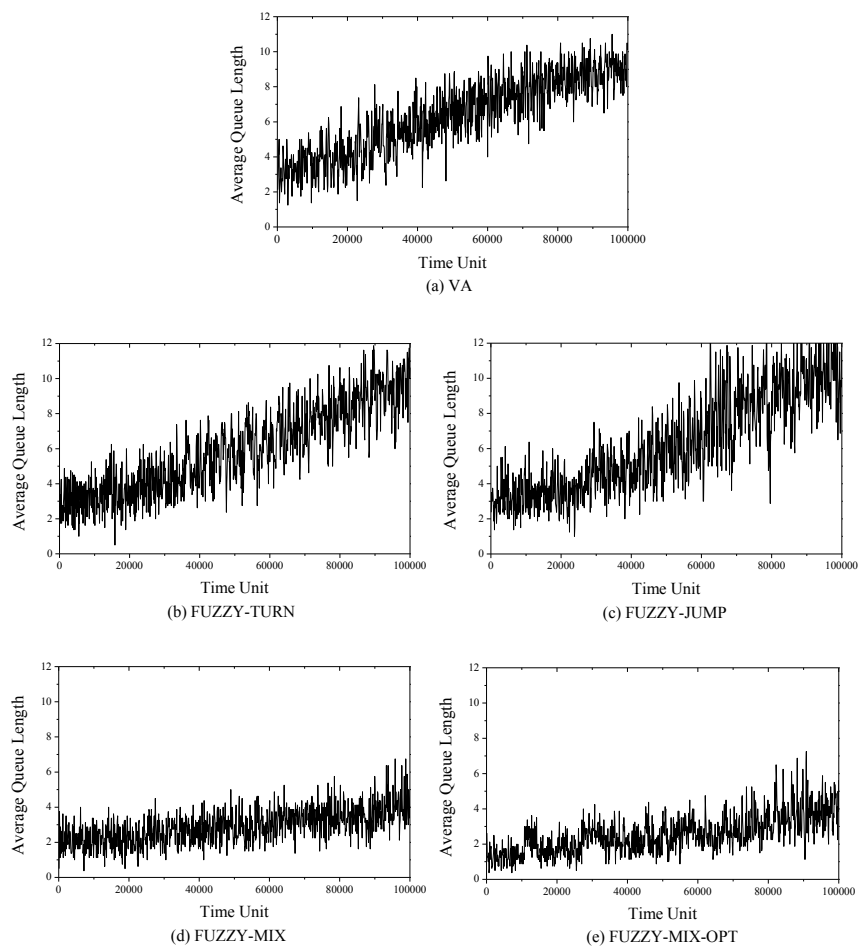

Fig. 11. Average queue length during the simulation of condition C16.

FUZZY-MIX-OPT controller for controlling the traffic flows of the roundabout is developed.

Simulations are done on a roundabout with four approaches and two circulatory lanes. Simulation results show that the proposed FUZZY-MIX controller outperforms the vehicle actuated method, the FUZZY-TURN controller, and the FUZZY-JUMP controller in terms of number of undetected vehicles and average vehicle delays. In addition, the FUZZY-MIX-OPT controller further improves the traffic efficiency obtained by FUZZY-MIX. The proposed FUZZY-MIX-OPT controller is very effective and can be regarded as a suggested scheme for the signal control of modern roundabouts.